# Large Scale Many-Objective Optimization Driven by Distributional Adversarial Networks

Zhenyu Liang, Yunfan Li, Zhongwei Wan

*Abstract*— Estimation of distribution algorithms (EDA) as one of the EAs are stochastic optimization problem which establishes a probability model to describe the distribution of solutions and randomly samples the probability model to create offspring and optimize model and population. Reference Vector Guided Evolutionary (RVEA) based on EDA framework, having better performance to solve MaOPs. Besides, using the generative adversarial networks to generate offspring solutions is also a state-of-art thoughts in EAs instead of crossover and mutation. In this paper, we will propose a novel algorithm based on RVEA[1] framework and using Distributional Adversarial Networks (DAN) [2]to generate new offspring. DAN uses a new distributional framework for adversarial training of neural network and operates on genuine samples rather than a single points because the framework also leads to more stable training and extraordinarily better mode coverage compared to single-point-sample methods. Thereby, DAN can quickly generate offspring with high convergence regarding to same distribution of data. In addition, we also use Large-Scale Multi-Objective Optimization Based on A Competitive Swarm Optimizer (LMOCSO)[3] to adopts a new two-stage strategy to update position in order to significantly increase the search efficiency to find optimal solutions in huge decision space. The propose new algorithm will be tested on 9 benchmark problems in Large scale multi-objective problems (LSMOP). To measure the performance, we will compare our proposal algorithm with some state-of-art EAs e.g., RM-MEDA[4], MO-CMA[10] and NSGA-II.

## I. Introduction

MULTIOBJETIVE optimal problems (MOPs) are common optimal problems in real world, and there already are several powerful estimation distribution algorithms (EDA) can deal with some optimal problems. But those algorithms like RM-MEDA [4] have good result only on such optimal problems that have objectives less than 3. In our project, we try to improve EDA to make it powerful on MaOPs.

### A. Multi-objective and Many-Objective

Since in real world, problems always depend on many factors, this is why we try to find a algorithm to deal with MOPs.

The definition of MOPs can be represented by this format:

$$\min y = F(\mathbf{x}) = (f_1(\mathbf{x}), f_2(\mathbf{x}),..., f_M(\mathbf{x}))$$
$$s.t. \quad \mathbf{x} \in X \tag{1}$$

where $X \in R^n$ is the decision space on Euclidean space with $\mathbf{x} = (x_1, x_2, x_3, \ldots, x_n) \in X$ is a vector and M is the number of objectives on problem $F$. For MOPs, M always is 2 or 3, and for those problems have objectives more than 3, called many-objective problems (MaOPs).

### B. Estimation Distribution Algorithm

In fact, EDA is a stochastic optimization algorithm. It has population which is a set of several solution for MOPs and offspring which is new solution created by algorithm and selection which is used to select better solution. But the main difference between EDA and normal EA is that EDA use a possibility model to create offspring and optimize model and population. The common framework of EDA is shown below:

**Step 0) Initialization:** Set generation t as 0. Generate an initial population $P(0)$. Use objective function to evaluate each individual $x$ and set these vectors as $F(0)$. Create an initial model as $M(0)$.

**Step 1) Stopping Condition:** If stopping condition is met, stop and return population $P(t)$. Then use $f$ to calculate $F(t)$, and $F(t)$ is the set that approach to Pareto front.

**Step 2) Sampling:** Use model $M(t)$ to generate new offspring $Q(t)$ and combine population $P(t)$ and offspring $Q(t)$ as a new large population $P(t)$.

**Step 3) Selecting:** Use selecting method to get better individual in population that approach to Pareto front and set these better individuals as new population $P(t+1)$.

**Step 4) Updating:** Update model by $P(t+1)$ and $M(t)$, then set new model as $M(t+1)$.

**Step 5) Setting generation:** Set generation t as t+1 and go to **Step 1)**.

## II. BACKGROUND

*A. DAN*

DAN [2] is an interesting variant of generative adversarial network (GAN) It also follows the design of GAN which uses a neural net called generator (G) and another neural net called discrimination (D). Besides, it introduces two aspects called Deep Mean Encoder (DME) [2] and Two-sample Classifier ($M_{2s}$) to improve the efficiency of model.

The generator G is a part of neural network to generate data G(z) that whose distribution close to distribution $P_x$ from a noise z. The discriminator D is another part of neural network to estimate the confidence of G(z) to contribution $P_x$. DME η is a encoder that use mean to estimate the distance between distributions which services for $M_{2s}$, which has theoretical guarantee from Maximum Mean Discrepancy (MMD) [5] and followed the form

$$\eta(\mathbf{P}) = \mathbf{E}_{x \sim \mathbf{P}}[\varphi(x)] \quad (2)$$

where φ is parameters in model trained by neural network and x is the input variable subjects to a distribution.

$M_{2s}$ is kernel idea of DAN, which in fact can be thought as a classifier rather than discriminator. It uses cross entropy to estimate the difference of two distributions. To improve the efficiency of discrimination, it consider cross the variables of distributions to sufficiently utilize the information in data. It divided real data X to first part $X_1$ and second part $X_2$ and generated Y to first part $Y_1$ and second part $Y_2$. Then let first part data $P_1$ minus second part data $P_2$, and estimate the cross entropy of absolute result of it, which is shown as eq. (4). As above, discriminator $D_{2S}$ will discriminate four variable group on $M_{2s}$, which follows such objective function.

$d_{2s}(\mathbf{P_1}, \mathbf{P_2}) = \log(M_{2s}(X_1, X_2)) + \log(M_{2s}(Y_1, Y_2)) + (1 - \log(M_{2s}(X_1, Y_2))) + (1 - \log(M_{2s}(Y_1, X_2)))$ (3)

$M_{2s}(X_1, X_2) = D_{2S}(|\eta(X_1) - \eta(X_2)|)$ (4)

Where $X = \{x_i\}_{i=1}^{n} \sim \mathbf{P_1}$, $Y = \{y_i\}_{i=1}^{n} \sim \mathbf{P_2}$, $X_1 := \{x_i\}_{i=1}^{\frac{n}{2}}$, $X_2 := \{x_i\}_{\frac{n}{2}+1}^{n}$, $Y_1 := \{y_i\}_{i=1}^{\frac{n}{2}}$, $Y_2 := \{y_i\}_{\frac{n}{2}+1}^{n}$.

DAN is still a generative adversarial network, and its steps as follows. It sets input data as real data and to training generator and discriminator. Generator is used to approach noise to real data, and discriminator is used to evaluate the correctness of data that come from generator. Every time discriminator have evaluate the correctness of data, it will send feedback to itself and generator, generator will try to cheat discriminator and discriminator want to evaluate data correctly. In this process two net will improve their performance until termination. During this process, two-sample classifier $M_{2s}$ will used to input as part of feedback , which is a way to estimate relevancy of real data contribution and generated data contribution. $M_{2s}$ will be update in execution of model, but not every circle.

*B. LMOCSO*

LMOCSO is the algorithm with competitive swarm optimizer. It adopts a competitive mechanism to determine the particles to be updated. At the beginning, calculate the fitness of each particle using the following fitness function.

$$Fitness(\mathbf{p}) = \min_{\mathbf{q} \in P \setminus \{\mathbf{p}\}} \sqrt{\sum_{i=1}^{M}(\max\{0, f_i(\vec{q}) - f_i(\vec{p})\})^2} \quad (5)$$

where $f_i(\vec{p})$ denotes the i-th objective value of p and M denotes the number of objectives.

Then two particles are randomly picked up from the current population $P$. and the one with smaller fitness value $x_l$ is updated by learning from the other one $x_w$ by Eq. (6).

$\vec{v_l}(t+1) = r_0\vec{v_l}(t) + r_1(\vec{x_w}(t) - \vec{x_l}(t)), \vec{x_l}(t+1) = \vec{x_l}(t) + \vec{v_l}(t+1) + r_0(\vec{v_l}(t+1) - \vec{v_l}(t))$ (6)

where $r_0$ and $r_1$ are uniformly randomly distributed values in [0, 1].

Afterwards, particle are mutated by polynomial mutation and put into the new population $P$.

*C. RVEA*

In RVEA, it uses reference vectors to select new population.

Reference vectors is unit vectors uniformly distributed inside the first quadrant [1]. It generated by norm. It can use a method called canonical simplex-lattice design [8] to create uniformly distributed point inside first quadrant, then use its norm can get uniformly distributed reference vectors. These reference vectors will refer to Pareto front later.

In selection part, there are three steps need to do. First, to translate objective value of individuals to first quadrant. Then divide those individuals into N partition, where N is the number of reference vectors. Lastly, use Angle-Penalized Distance (APD) Calculation [1] to calculate the APD between reference vectors and select the individual have minimal APD on its translated objective value for each reference vectors.

## III. PROPOSED ALGORITHM

### A. Main Framework

The main framework of our algorithm MOEA based on a competitive swarm optimizer and DAN (MOEA-CSOD) is shown below[1]:

**Step 0) Initialization:** Generate the initial population $P_0$ with $N$ randomized individuals and a set of unit reference vector $V_0 = \{\mathbf{v}_{0,1}, \mathbf{v}_{0,2}, ..., \mathbf{v}_{0,N}\}$.

**Step 1) Stopping Condition:** If stopping condition is met, stop and return population $P_t$. Then use $f$ to calculate $F_t$.

**Step 2) Offspring Creation:** Use DAN and LMOCSO to create the new offspring. Use probability to choose which method to create new individuals.

**Step 3) Selecting:** Use reference vector-guided selection to select N individuals.

**Step 4) Adaptation:** Adapt the reference vector.

**Step 5) Setting generation:** Set generation t as t+1 and go to **Step 1)**.

### B. Offspring Creation

In our paper, we generate offspring by by LMOCSO and DAN. We use a random method to get a probability number $\lambda$ which range from 0.2 to 0.8 to get decide percentage of new individuals from DAN and use LMOCSO to get rest.

At the part of LMOCSO, the details are in the background.

At the part of DAN, since we have a set of data about a population, we use it as a real data in our neural network model and use a Gaussian noise [6] to generator G.

When the DAN is training, the generator G will use a noise Z to generator new data G(Z), then DME encodes G(Z) and real data X to G(Z)' and X'. After that, divides G(Z)' and X' into G(Z$_1$)', G(Z$_2$)', X$_1'$, X$_1'$ into D$_{2S}$ to estimate and use its objective function to update the **M$_{2s}$**. In fact, the update of **M$_{2s}$** is always needed. Then, input data G(Z) to the discriminator D to judge the confidence of G(Z) to contribution **X**. After such steps, updates D with objective function of D and updates G with objective function of D and **M$_{2s}$**. The loss function of model is shown as

$$\min_{G} \max_{D,M_{2S}} V(G,D,M_{2S}) = \lambda_1 E_{x \sim p_x, z \sim P_z}[\log D(x) + \log(1 - D(G(z)))] + \lambda_2 d_{2s}(P_x, P_G) \quad (7)$$

When DAN is trained over, the generator G in DAN can be used to generate next generation with a Gaussian noise.

### C. Reference Vector-Guided Selection[1]

There are four steps in reference vector-guided selection: 1) objective value translation; 2) population partition; 3) APD calculation; and 4) the elitism selection.

1) *Objective Value Translation*: $F_t = \{\mathbf{f}_{t,1}, \mathbf{f}_{t,2}, ..., \mathbf{f}_{t,|P_t|}\}$ donates the objective values of the population. Objective values translate from $F_t$ to $F_t'$ via

$$\mathbf{f}_{t,i}' = \mathbf{f}_{t,i} - \mathbf{z}_t^{min} \quad (8)$$

, where i = 1, ... , |P$_t$|, $\mathbf{f}_{t,i}$ and $\mathbf{f}_{t,i}'$ are the objective vectors of individual i before and after the translation. Also $\mathbf{z}_t^{min} = (z_{t,1}^{min}, z_{t,2}^{min}, ..., z_{t,m}^{min})$ represents the minimal objective values calculated from F$_t$.

2) *Population Partition*:

After the translation, the population will be partitioned into N subpopulation by associating each individual with its closest reference vector. To get the closest reference vector, we can calculate the angle between reference vector and objective vector:

$$\cos\theta_{t,i,j} = \frac{\mathbf{f}_{t,i}' \cdot \mathbf{v}_{t,j}}{\|\mathbf{f}_{t,i}'\|} \quad (9)$$

where $\theta_{t,i,j}$ represents the angle between objective vector $\mathbf{f}_{t,i}'$ and reference vector $\mathbf{v}_{t,j}$.

Then choose the maximal cosine to partition:

$$\bar{P}_{t,k} = \{I_{t,i} | k = \arg\max \cos\theta_{t,i,j}\} \quad (10)$$

where $I_{t,i}$ denotes the *i*th individual in *P*t.

3) *Angle-Penalized Distance calculation*: Calculate APD by:

$$d_{t,i,j} = (1 + P(\theta_{t,i,j})) \cdot \|\mathbf{f}_{t,i}'\| \quad (11)$$

where $P(\theta_{t,i,j})$ is a penalty function related to $\theta_{t,i,j}$.

$$P(\theta_{t,i,j}) = M \cdot \left(\frac{t}{t_{max}}\right)^\alpha \cdot \frac{\theta_{t,i,j}}{\gamma_{v_{t,j}}} \quad (12)$$

$$\gamma_{v_{t,j}} = \min_{i \in \{i,...N\}, i \neq j} \langle v_{t,i}, v_{t,j} \rangle \quad (13)$$

where M is the number of objectives, N is the number of reference vectors, $t_{max}$ is the predefined maximal number of generations, $\gamma_{v_{t,j}}$ is the smallest angle value between reference

After selection, it is necessary to adapt reference vectors to make sure its reference function.

vector and the other reference vectors in the current generation, and α is a user defined parameter controlling the rate of change of $P(\theta_{t,i,j})$.

4) *Elitism Selection*:

Select the smallest APD of population as the elite population.

D. *Reference Vector Adaptation[1]*

Instead of normalizing the objectives, the algorithm adapt the reference vectors according to the ranges of the objective values in the following manner:

$$\mathbf{v}_{t+1,i} = \frac{\mathbf{v}_{0,i} \circ (\mathbf{z}_{t+1}^{max} - \mathbf{z}_{t+1}^{min})}{\|\mathbf{v}_{0,i} \circ (\mathbf{z}_{t+1}^{max} - \mathbf{z}_{t+1}^{min})\|} \quad (14)$$

$\mathbf{v}_{t+1,i}$ denotes the ith adapted reference vector for the next generation $t + 1$, $\mathbf{v}_{0,i}$ denotes the *i*th uniformly distributed reference vector of initial stage and $\mathbf{z}_{t+1}^{max}$ and $\mathbf{z}_{t+1}^{min}$ denote the maximum and minimum values of each objective function in the $t + 1$ generation.

IV. COMPARATIVE STUDIES

In this part of the experiment, we will use LSMOP1-LSMOP9 test suites to evaluate the performance of the four algorithms MOEA-CSOD, RM-MEDA, MO-CMA and NSGA-II. After that, we will get the IGD value from the result of the algorithm. The brief descriptions between the LSMOP1-9 suites are as follows.

$$\begin{cases} f_1(\mathbf{x}) = h_1(\mathbf{x}^f)(1 + g_1(\mathbf{x}^s)) \\ f_2(\mathbf{x}) = h_2(\mathbf{x}^f)(1 + g_2(\mathbf{x}^s)) \\ \dots \\ f_M(\mathbf{x}) = h_M(\mathbf{x}^f)(1 + g_M(\mathbf{x}^s)) \end{cases} \quad (15)$$

$$L_1(\mathbf{x}^s) = \left(1 + \frac{i}{|\mathbf{x}^s|}\right) \times (x_i^s - l_i) - x_1^f \times (u_i - l_i) \quad (16)$$

$$L_2(\mathbf{x}^s) = \left(1 + \cos\left(0.5\pi \frac{i}{|\mathbf{x}^s|}\right)\right) \times (x_i^s - l_i) - x_1^f \times (u_i - l_i) \quad (17)$$

Most algorithm studies on multi-objective optimization are limited to small-scale problems, but the fact is that real world multiple objective problems may contains a large number of decision variables and more than 3 objects. Large Scale Multi-Objective Problems (LSMOP) is a new proposal challenge MOPs consider mixed separability between decision variables nonuniform correlation between decision variables and objective functions. From Eq.(15) we can realize the basic form

TABLE I

IGD values of MOEA-CSOD, NSGA-II, RM-MEDA and MOCMA on 3-objective LSMOP1-LSMOP9, what the best result on each test instance is shown in a bold font.

| Problem | N. | D. | MOEA-CSOD | NSGA-II | | RM-MEDA | | MOCMA | |
|---|---|---|---|---|---|---|---|---|---|
| LSMOP1 | 105 | 300 | 8.8242e-01 | **8.8020e-01** | ≈ | 8.9974e+00 | - | 6.5324e+00 | - |
| LSMOP2 | 105 | 300 | 3.6731e-01 | **3.6259e-01** | ≈ | 9.3889e+00 | - | 6.7875e+00 | - |
| LSMOP3 | 105 | 300 | **9.1850e-01** | 9.7026e-01 | - | 8.8690e+00 | - | 6.7024e+00 | - |
| LSMOP4 | 105 | 300 | 6.7715e-01 | **6.7094e-01** | ≈ | 8.2954e+00 | - | 6.8129e+00 | - |
| LSMOP5 | 105 | 300 | 7.3072e-01 | 2.3330e+00 | - | 9.9355e+00 | - | 7.1733e+00 | - |
| LSMOP6 | 105 | 300 | 7.8590e+00 | **4.8614e+00** | + | 8.9211e+00 | - | 7.5286e+01 | + |
| LSMOP7 | 105 | 300 | 3.7729e+00 | **2.6937e+00** | + | 8.3872e+00 | - | 7.4121e+00 | - |
| LSMOP8 | 105 | 300 | **5.2350e-01** | 2.4226e+00 | - | 8.1250e+00 | - | 7.3251e+00 | - |
| LSMOP9 | 105 | 300 | 2.9464e+00 | **2.0100e+00** | + | 9.1896e+00 | - | 7.2706e+00 | - |
| + / - / ≈ | | | | 3/3/3 | | 0/9/0 | | 1/8/0 | |

'+', '−' and '≈' indicate that the result is significantly better, significantly worse and statistically similar to that of MOEA-CSOD, respectively.

TABLE II

IGD values of MOEA-CSOD, NSGA-II, RM-MEDA and MOCMA on 6-objective LSMOP1-LSMOP9, what the best result on each test instance is shown in a bold font.

| Problem | N. | D. | MOEA-CSOD | NSGA-II | | RMMEDA | | MOCMA | |
|---|---|---|---|---|---|---|---|---|---|
| LSMOP1 | 132 | 600 | 9.5859e-01 | **9.5720e-01** | ≈ | 4.6127e+01 | - | 5.9166e+10 | - |
| LSMOP2 | 132 | 600 | **4.0622e-01** | 4.0626e-01 | ≈ | 5.2637e+01 | - | 5.9796e+01 | - |
| LSMOP3 | 132 | 600 | 3.6318e+00 | **1.8878e+00** | + | 4.9699e+01 | - | 6.0508e+01 | - |
| LSMOP4 | 132 | 600 | 4.9019e-01 | **4.8146e-01** | + | 4.8855e+01 | - | 6.0911e+01 | - |
| LSMOP5 | 132 | 600 | **8.3704e-01** | 2.3847e+00 | - | 5.5975e+01 | - | 5.9916e+01 | - |
| LSMOP6 | 132 | 600 | **2.2061e+00** | 3.0912e+01 | - | 5.0917e+01 | - | 6.1023e+01 | - |
| LSMOP7 | 132 | 600 | **2.6629e+00** | 8.4705e+00 | - | 6.0510e+01 | - | 5.1974e+01 | - |
| LSMOP8 | 132 | 600 | **7.1447e-01** | 2.0581e+00 | - | 6.4456e+01 | - | 6.3983e+01 | - |
| LSMOP9 | 132 | 600 | **4.6384e+00** | 6.4775e+00 | - | 4.7366e+01 | - | 6.0259e+01 | - |
| + / - / ≈ | | | | 2/5/2 | | 0/9/0 | | 0/9/0 | |

'+', '−' and '≈' indicate that the result is significantly better, significantly worse and statistically similar to that of MOEA-CSOD, respectively.

TABLE III

IGD values of MOEA-CSOD, NSGA-II, RM-MEDA and MOCMA on 8-objective LSMOP1-LSMOP9, what the best result on each test instance is shown in a bold font.

| Problem | N. | D. | MOEA-CSOD | NSGA-II | | RMMEDA | | MOCMA | |
|---|---|---|---|---|---|---|---|---|---|
| LSMOP1 | 156 | 800 | 1.0019e+00 | **9.7266e-01** | + | 1.2084e+02 | - | 1.1636e+01 | - |
| LSMOP2 | 156 | 800 | 4.4744e-01 | 4.3077e-01 | + | 1.8860e+02 | - | **3.5037e-01** | + |
| LSMOP3 | 156 | 800 | **1.8451e+00** | 2.0146e+00 | - | 9.6733e+01 | - | 2.7557e+01 | - |
| LSMOP4 | 156 | 800 | 5.1614e-01 | 5.3520e-01 | - | 1.0608e+02 | - | **3.8487e-01** | + |
| LSMOP5 | 156 | 800 | **9.4178e-01** | 3.7212e+00 | - | 1.1214e+02 | - | 1.6086e+01 | - |
| LSMOP6 | 156 | 800 | 4.8915e+00 | 1.9432e+01 | - | 1.2558e+02 | - | **1.8805e+00** | + |
| LSMOP7 | 156 | 800 | **2.9180e+00** | 8.4362e+00 | - | 1.3795e+02 | - | 5.1819e+04 | - |
| LSMOP8 | 156 | 800 | **8.4175e-01** | 1.4497e+00 | - | 1.1941e+02 | - | 1.4413e+01 | - |
| LSMOP9 | 156 | 800 | **6.5945e+00** | 1.2537e+01 | - | 9.7661e+01 | - | 7.1399e+02 | - |
| + / - / ≈ | | | | 2/7/0 | | 0/9/0 | | 3/6/0 | |

'+', '−' and '≈' indicate that the result is significantly better, significantly worse and statistically similar to that of MOEA-CSOD, respectively.

TABLE IV

IGD values of MOEA-CSOD, NSGA-II, RM-MEDA and MOCMA on 10-objective LSMOP1-LSMOP9, what the best result on each test instance is shown in a bold font.

| Problem | N. | D. | MOEA-CSOD | NSGA-II | | RMMEDA | | MOCMA | |
|---|---|---|---|---|---|---|---|---|---|
| LSMOP1 | 275 | 1000 | 9.8445e-01 | **9.8156e-01** | ≈ | 2.3827e+02 | - | 1.1761e+01 | - |
| LSMOP2 | 275 | 1000 | 4.4727e-01 | 4.4642e-01 | ≈ | 3.4939e+02 | - | **3.3089e-01** | + |
| LSMOP3 | 275 | 1000 | 2.2491e+00 | **7.1451e-01** | + | 1.3016e+01 | - | 2.1722e+01 | - |
| LSMOP4 | 275 | 1000 | 4.7954e-01 | 4.4294e-01 | + | 3.1241e+02 | - | **3.8113e-01** | + |
| LSMOP5 | 275 | 1000 | **9.8300e-01** | 4.7387e+00 | - | 3.3594e+02 | - | 1.8965e+01 | - |
| LSMOP6 | 275 | 1000 | 1.7767e+00 | 2.7920e+01 | - | 2.7164e+02 | - | **1.5484e+00** | + |
| LSMOP7 | 275 | 1000 | **3.5608e+00** | 8.3191e+01 | - | 2.5041e+02 | - | 5.6880e+04 | - |

| | | | | | | | | |
|---|---|---|---|---|---|---|---|---|
| LSMOP8 | 275 | 1000 | **1.5662e+00** | 5.8376e+00 | - | 2.6937e+02 | - | 1.2908e+01 | - |
| LSMOP9 | 275 | 1000 | **1.7890e+01** | 2.2723e+01 | - | 2.1703e+02 | - | 1.1726e+03 | - |
| +/-/≈ | | | | 2/5/2 | | 0/9/0 | | 3/6/0 | |

'+', '−' and '≈' indicate that the result is significantly better, significantly worse and statistically similar to that of MOEA-CSOD, respectively.

of proposed LSMOP problem, h(x) define the shape of PF, g(x) defines the fitness landscape, known as landscape functions hereafter. There are 9 proposed problem from LSMOP1 to LSMOP2 have been instantiated and based on generic principle. Among these LSMOPs, LSMOP1 to LSMOP4 have a linear pareto front (PF), and LSMOP5-LSMOP8 have a non-linear PF, and LSMOP9 has a disconnected PF. Besides, the LSMOP1-LSMOP4 are designed to have linear variable linkage but LSMOP5-LSMOP9 is nonlinear-linkage. The problems also contains separable correlation, overlapped correlation and full correlation.

TABLE I-4 is the IGD values of the MOEA-CSOD, NSGA-II, RM-MEDA and MOCMA algorithm on 3, 6, 8 ,10-objective LSMOP1-LSMOP9. The number of population is N and the dimension of individual is D. It will evolve 50 generations.

According to the TABLE I-IV, MOEA-CSOD show good performance on 6,8,10-objective LSMOP5-LSMOP9, which means our algorithm can better solve the large scale and difficult problem. However, when solve 3-objective LSMOP, it is similar to NSGA-II.

## V. CONCLUSION

According to the table, we can find that MOEA-CSOD have better performance in LSMOP. However, there are some problem in our algorithm. The first one is the final generation has a small number of individual and the second is DAN may have the problem of overfitting. Also, MOEA-CSOD needs to be further improved on small scale MOPs(i.e. LSMOP1 and LSMOP4 ) All in all, we will improve our algorithm in the coming days.